\documentclass[10pt,a4paper,twocolumn]{article}

\usepackage[a4paper,margin=0.85in,columnsep=0.32in]{geometry}
\usepackage{microtype}
\usepackage{setspace}
\usepackage{amsmath,amssymb,amsthm}
\usepackage{booktabs}
\usepackage{multirow}
\usepackage{siunitx}
\usepackage{graphicx}
\usepackage{caption}
\usepackage{subcaption}
\usepackage{placeins}
\usepackage{float}
\graphicspath{{figures/}}

\usepackage[round,sort&compress]{natbib}
\usepackage[dvipsnames]{xcolor}
\usepackage[colorlinks=true,linkcolor=blue!50!black,citecolor=blue!50!black,urlcolor=blue!50!black]{hyperref}
\usepackage{cleveref}
\usepackage{xurl}

\usepackage{verbatim}

\title{Brand Recall and Category Accuracy in Language Models:\\
An Audit of 1,000 B2B SaaS Brands}

\author{%
  Ben Moore\\
  Discovered Labs\\
  \url{https://discoveredlabs.com}\\
  \texttt{ben@discoveredlabs.com}
}

\date{November 23, 2026}
\hypersetup{pdftitle={Brand Recall and Category Accuracy in Language Models: An Audit of 1,000 B2B SaaS Brands},pdfauthor={Ben Moore}}

\begin{document}
\maketitle

\begin{abstract}
A language model can name a brand without classifying it correctly. This study
measures these outcomes separately for 1,000 B2B SaaS brands across 333 product
categories. On Gemma 4 12B, with no external retrieval, category prompts elicit
510 brands (51.0\%). A separate five-candidate category probe ranks the reference
label first for 787 brands, a distractor first for 194, and yields little
separation for 19. Among the 981 separated rankings, the error rate is 19.8\%.
It falls to 3.5\% for the 547 brands with a winner-to-runner-up log-probability
margin above 0.6. These are within-cohort measurements, not calibrated
confidence estimates. Exploratory regressions associate technical coverage
with recall and an owned-properties composite with category accuracy, but the
composites use different adjustment sets and do not establish causal effects.
Across four Gemma generations, measured recall is 39.8\%, 44.3\%, 51.1\% and
51.0\%. A separate representation-based sentiment measure has little linear
association with customer ratings ($r=0.03$, $n=921$). The findings support
reporting recall and category accuracy separately. Failure to elicit a brand
does not establish its absence from model parameters or training data.
\end{abstract}


\section{Introduction}
\label{sec:intro}

When a language model is used for product research, two questions matter: does
it name a relevant company, and does it identify what that company does?
Measuring mentions alone cannot distinguish these outcomes. Retrieval-enabled
assistants add another source of variation because their answers may depend on
newly retrieved documents as well as model parameters.

This study separates sampled recall from category ranking on an open-weight
model without external retrieval. Recall is measured from generated
completions; category accuracy is measured by scoring fixed candidate
sentences. Neither measurement directly observes the training corpus or proves
that a particular fact is absent from the model.

The study makes three contributions:
\begin{itemize}\itemsep2pt
\item A retrieval-free audit of sampled recall and category ranking for
      1,000 B2B SaaS brands across 333 categories.
\item A descriptive comparison of recall across four Gemma generations and
      an analysis of category error rates by likelihood margin.
\item Exploratory associations between public brand characteristics and the
      two outcomes, with secondary commercial-fact, sentiment and
      membership-detector analyses.
\end{itemize}
The main result is that recall and category accuracy differ in their observed
rates and associated variables. This is a measurement framework, not a causal
prescription for changing model outputs.

\section{Related work}
\label{sec:related}

\paragraph{Factual probing.} Cloze-style probes evaluate whether language
models recover known relations \citep{petroni2019lama}. Work on locating and
editing factual associations studies how such behavior relates to model
parameters \citep{meng2022rome}. Our category probe ranks candidate statements.
Its outcome depends on wording and distractors, so it is evidence about a
specified probe rather than a complete inventory of stored knowledge.

\paragraph{Training-data inference.} Extracting memorized text
\citep{carlini2021extracting} and detecting likely training examples are related
but distinct tasks. Min-K\% and Min-K\%++ use token probabilities to detect
statistical differences between member and non-member text
\citep{shi2024mink,zhang2025minkpp}. We use a reference-normalized detector for
an exploratory passage survey. Calibration on one corpus does not establish
membership for a model with an undisclosed training set.

\paragraph{Representations and brand measurement.} Representation engineering
studies directions in hidden states associated with semantic attributes
\citep{zou2023representation}. Our sentiment analysis uses contrast-derived
directions, with comparisons made within each dimension. This complements
output-based brand measurement, but does not establish that the directions
measure customer experience or that an assistant will express the same scores.

\section{Data}
\label{sec:data}

\paragraph{Cohort.} The study contains 1,000 B2B SaaS brands spanning 333
product categories, assembled from public information. Each brand has a single
reference category used in the category probe. This is an English-language,
non-random cohort. The term public refers to information availability, not to
stock-market listing. A single reference category can understate the range of
products sold by a company.

\paragraph{Models.} The primary run is recorded as Gemma 4 12B. The
cross-generation comparison adds Gemma 7B, Gemma 2 9B and Gemma 3 12B using the
same cohort and recall protocol. Separate API runs cover Claude Haiku 4.5,
Gemini 3.5 Flash Lite and GPT-5.6 Luna. Their results are reported descriptively
because their category tasks and sampling budgets differ from the Gemma runs.

\paragraph{Brand characteristics.} Public measurements include mention counts
and audience sizes; the share of coverage that states a brand's category;
owned-site documentation, help centres and commercial information; and name
and category characteristics. The per-variable analysis groups these into
coverage volume, coverage language, owned properties and namespace for
multiple-testing correction. These are analysis groupings, not claims that
all signals within a group measure one underlying construct.

\paragraph{Coverage.} The association dataset contains 998 brands. Complete-case
sample sizes vary by predictor: the saved per-variable fits use 178--885 brands
for recall and 174--869 for category accuracy. Composite fits use 855--998.
Missing surfaces and measurements may be informative rather than random.
Common Crawl host-graph rank and referring-domain counts serve as measures of
web prominence, with the adjustment exceptions stated below.

\section{Methods}
\label{sec:methods}

\subsection{Sampled category recall}
For each category, two unbranded prompts request examples of relevant software.
Six completions are sampled per prompt at temperature 0.7. A brand counts as
recalled if it occurs in any of its category's twelve completions. Matching
uses word boundaries, legal-suffix stripping and domain forms. This measures
elicitation under a finite sampling budget. Non-recall is not proof of absent
knowledge. An earlier substring matcher produced false positives and was
replaced before the reported analysis.
In a reseed comparison covering 321 brands in 111 categories, 42 brands
(13.1\%) changed recall status. This is a per-brand stability check, not a
confidence interval for the full-cohort recall rate.

\subsection{Category ranking}
For each brand, five sentences differ only in the stated product category:
one reference label and four distractors. Each sentence is scored by mean
token log-probability. The saved probe assigns one of three outcomes:
\begin{itemize}\itemsep2pt
\item \textbf{Consistent:} the reference category ranks first with sufficient
      score separation.
\item \textbf{Deviated:} a distractor ranks first with sufficient separation.
\item \textbf{Low separation:} the candidates are insufficiently separated
      (recorded as \texttt{void} in the output).
\end{itemize}
These are ranking outcomes, not observations of an assistant choosing to
answer or abstain.

The decision margin is the score difference between the highest-ranked and
second-ranked sentences. Distractors are matched to the reference label's
word count, with character length as a tiebreak. Right padding and attention
masking prevent padding from changing the score. Both label-length matching
and padding correction materially changed the results; the appendix reports
the sensitivity. Reported margin thresholds are evaluated on the same cohort
and have not been validated as confidence thresholds on independent brands.

\subsection{Commercial facts}
Free-tier availability, published pricing, self-serve signup and integration
partnerships are evaluated using positive-minus-negative log-probability
contrasts. Wording is held fixed while the brand varies. AUC measures whether
brands with a true claim tend to score above brands with a false claim.
Integration pairs use real cohort brands on both sides. Low AUC indicates weak
discrimination by these probes, not proof that the model contains no relevant
information.

\subsection{Representation-based sentiment}
The sentiment instrument uses twelve directions derived from antonym contrasts.
Brand representations are mean-pooled over brand-name tokens located by
character offsets at layer $-8$. Each dimension uses its own prompt frame.
Comparisons are therefore within a dimension, or use dimension-specific cohort
$z$-scores. These measurements are neither customer-satisfaction scores nor
probabilities of a generated answer expressing a sentiment.

\subsection{Exploratory associations}
Separate logistic regressions relate each brand characteristic to recall and
to accuracy among brands with separated category rankings. Per-variable models
adjust for crawl-graph rank and referring domains; accuracy models also include
the reference label's word-count rank in the original candidate set. This
control was retained from the original probes, not recomputed from the
length-matched candidates. The saved
analysis contains 87 recall fits and 86 accuracy fits. Benjamini--Hochberg
correction is applied within variable family and outcome
\citep{benjamini1995fdr}, not across the entire study.

The composite analysis fits six channel measures and two prominence baselines
against both outcomes, with correction across all 16 tests. Components are
standardized and averaged, requiring two observed components except for the
single-component category-explanation measure. Technical, explanation, review,
news and social composites adjust for crawl rank and referring domains. The
owned-properties composite includes crawl rank as a component and adjusts only
for referring domains. The two prominence baselines are unadjusted. Composite
accuracy fits do not include the label-rank control. Consequently the composite
coefficients are not comparisons under a common adjustment set.

Effects are reported in outcome percentage points per predictor standard
deviation. The implementation averages the logistic derivative and multiplies
it by the fitted coefficient. Displayed interval endpoints use the same average
derivative to scale coefficient confidence limits; they are approximate and do
not propagate uncertainty in that derivative. Complete-case fitting,
correlated predictors and unmeasured confounding limit interpretation. No
variable was experimentally manipulated.

\section{Results}
\label{sec:results}

\subsection{Recall covers 51\% of the cohort}
\begin{figure*}[t]
\centering
\includegraphics[width=\textwidth]{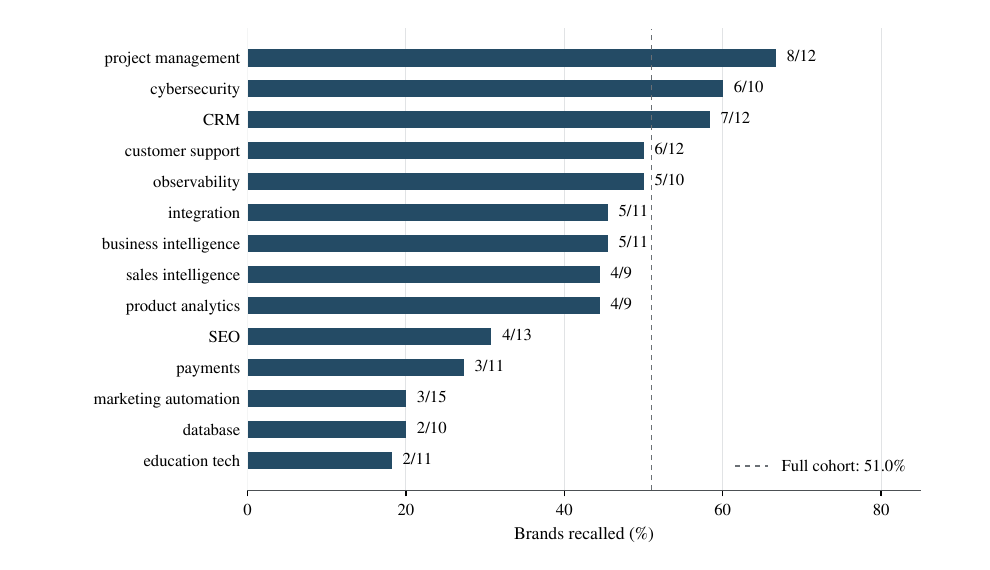}
\caption{Sampled recall in the 14 largest categories, selected by cohort size. Labels report recalled brands over category totals. The dashed line is the full-cohort rate (510/1,000). Small category samples make these descriptive comparisons.}
\label{fig:recall-categories}
\end{figure*}

Gemma 4 12B names 510 of 1,000 brands in the sampled, unbranded category
completions. The remaining 490 are not recalled under this protocol; this
does not establish that information about them is absent from the model or
its training data. Recall varies by category, including 8 of 12 project
management brands and 2 of 11 education technology brands
(\Cref{fig:recall-categories}).

\Cref{tab:generations} reports recall of 39.8\%, 44.3\%, 51.1\% and 51.0\%
across four Gemma generations. These are descriptive comparisons under the
same sampling protocol. Model size and training differ across generations,
and the small difference between Gemma 3 and Gemma 4 does not establish a
general plateau in brand knowledge.

\subsection{Category accuracy is 78.7\%}
In the five-candidate category probe, the designated true category ranks
first for 787 brands, a distractor ranks first for 194, and 19 have
insufficient score separation (\Cref{tab:outcomes}). The error rate is
19.4\% of the full cohort, or 19.8\% of the 981 cases with sufficient
separation. This denominator includes both recalled and unrecalled brands.
The low-separation classification is an analyst-defined probe outcome,
not a model-generated abstention.

\begin{figure}[htbp]
\centering
\includegraphics[width=\columnwidth]{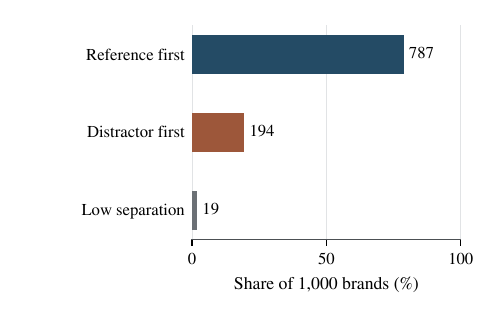}
\caption{Corrected category-ranking outcomes for all 1,000 brands. Low separation is an analyst-defined probe outcome, not an explicit model abstention.}
\label{fig:category-outcomes}
\end{figure}

\begin{figure*}[t]
\centering
\includegraphics[width=\textwidth]{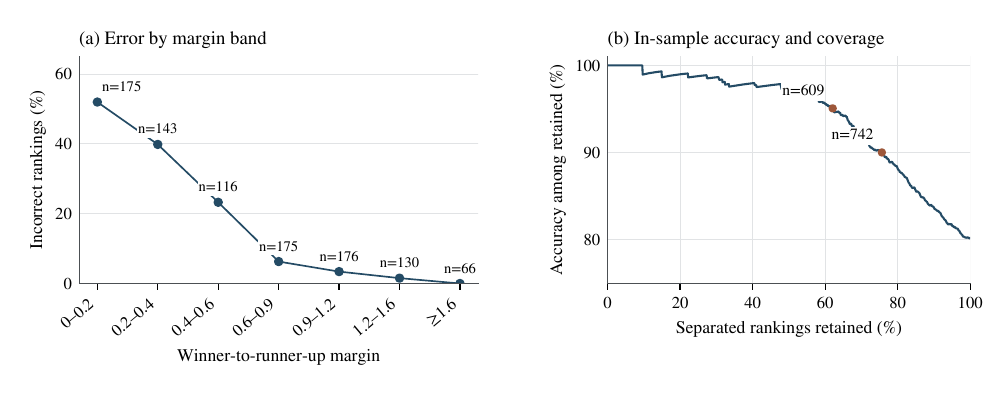}
\caption{Category-ranking error and coverage among 981 separated rankings. (a) Error rate in each margin band, with band sample sizes. (b) Accuracy when retaining brands above successive observed margin thresholds; tied margins are included together. Both panels describe the same labelled cohort and are not held-out calibration estimates.}
\label{fig:margin-accuracy}
\end{figure*}

Median winner-to-runner-up margins are 0.83 for correct rankings and 0.21
for incorrect rankings. Error rates decrease across the margin bands in
\Cref{fig:margin-accuracy} and \Cref{tab:margin}: the two lowest bands have error rates of 52.0\% and
39.9\%, while the 547 brands above 0.6 have a combined error rate of 3.5\%.
These are retrospective results on the same cohort. The margin has not
been calibrated on held-out brands and is not a measure of the confidence
expressed in a generated answer.

\subsection{Commercial-fact probes discriminate weakly}
Integration probes separate true from false brand pairs at AUC 0.614 over
2,400 probes. Free-tier, published-pricing and self-serve probes yield AUCs
of 0.547, 0.526 and 0.450 respectively (\Cref{tab:facttypes}). These
point estimates indicate limited discrimination under the tested wording;
they do not show that the facts are absent from the weights. Five-way
category accuracy and binary-probe AUC use different scales and should not
be compared directly.

\begin{figure}[htbp]
\centering
\includegraphics[width=\columnwidth]{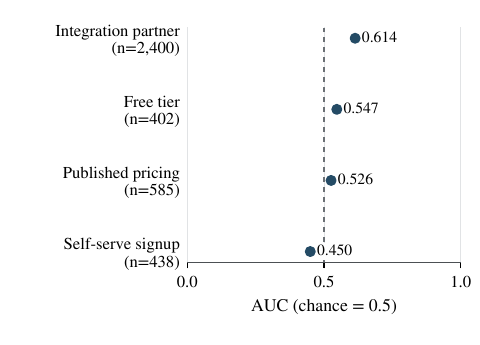}
\caption{Binary commercial-fact probe AUCs, with sample sizes. The dashed line marks chance discrimination (0.5). Category accuracy uses a different metric and is reported separately. Only point estimates are shown; no uncertainty intervals were stored for these summaries.}
\label{fig:fact-types}
\end{figure}

\subsection{Recall and accuracy have different associations}
The technical-coverage composite is associated with 5.37 percentage points
more recall per standard deviation ($p_{adj}=0.006$), after adjustment for
crawl-graph rank and referring domains. Its association with category
accuracy is $-2.77$ points ($p_{adj}=0.047$).

\begin{figure*}[t]
\centering
\includegraphics[width=\textwidth]{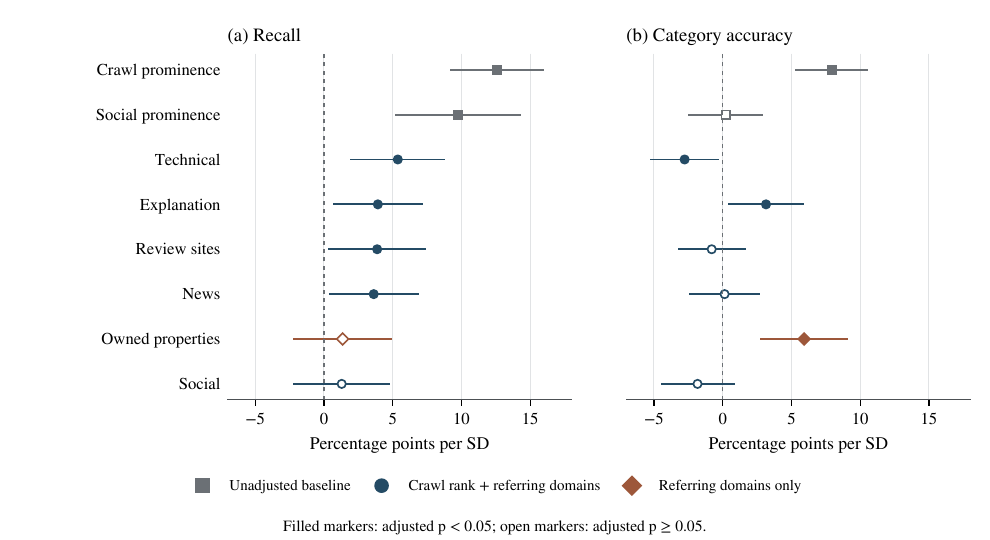}
\caption{Composite associations with recall and category accuracy. Points show percentage-point changes per predictor standard deviation; bars are the approximate 95\% intervals described in Methods. Shape distinguishes the unadjusted prominence baselines and the owned-properties adjustment exception. Open markers denote adjusted $p\geq0.05$ across 16 tests. All fitted composites are shown; their adjustment sets differ.}
\label{fig:composites}
\end{figure*}

The owned-properties composite is associated with 5.92 points more category
accuracy ($p_{adj}<0.001$); its recall association is 1.36 points and is not
statistically significant. This composite includes crawl-graph rank and
adjusts only for referring domains. Its association therefore cannot be
attributed to owned content independently of crawl prominence. The two
prominence baselines are unadjusted. \Cref{tab:composites} reports all eight
composites, with correction across their sixteen outcome tests.
\Cref{fig:composites} compares the corresponding estimates.

The explanation composite, based on category co-occurrence, is associated
with both outcomes in the composite analysis. Its recall association does
not survive the separate within-family correction in the individual-variable
analysis ($p_{adj}=0.077$). This sensitivity to the testing family limits
any claim that category co-occurrence is a robust independent predictor.
The individual-variable results are reported in \Cref{app:adjusted}.

\subsection{Sentiment scores do not track customer ratings}
For 921 brands with customer-rating data, the correlation with the model
sentiment measure is $r=0.030$ ($p=0.36$). The study therefore provides no
evidence that this score is a customer-satisfaction proxy. It does not
identify the source of the sentiment signal.

\begin{figure*}[t]
\centering
\includegraphics[width=\textwidth]{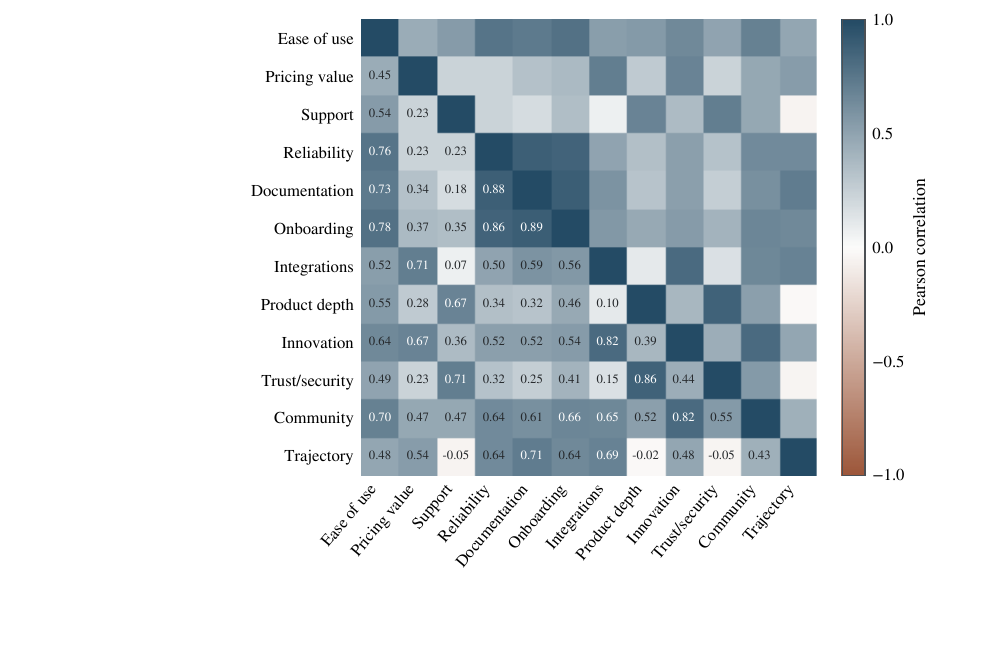}
\caption{Pearson correlations between the 12 sentiment dimensions across 1,000 brands with complete scores. The fixed colour scale spans $-1$ to $1$. This plot measures overlap within the sentiment instrument, not its correlation with customer ratings.}
\label{fig:sentiment-correlations}
\end{figure*}

The twelve dimensions overlap: their median pairwise correlation is 0.517,
and the first principal component accounts for 54.8\% of their variance.
Dimension-specific readings should consequently not be treated as twelve
independent measurements or compared on their raw scales.
\Cref{fig:sentiment-correlations} shows their pairwise correlations.

\subsection{Detector scores vary with passage format}
\begin{figure}[htbp]
\centering
\includegraphics[width=\columnwidth]{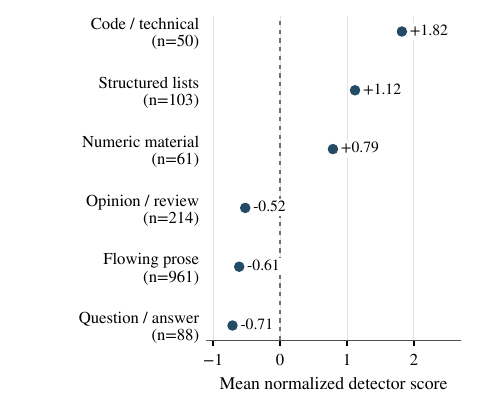}
\caption{Mean reference-normalised detector scores by passage format, with passage counts ($n=1,477$ in total). Scores are indirect detector measurements, not verified membership rates. Only means are available in the saved summary; uncertainty intervals are not shown.}
\label{fig:content-format}
\end{figure}

Across 1,477 passages, the reference-normalised detector has mean scores
of $+1.82$ for code and technical material, $+1.12$ for structured lists,
and $+0.79$ for numeric material. Personal opinion, flowing prose and
question-and-answer passages score $-0.52$, $-0.61$ and $-0.71$ respectively.
These are detector-score differences, not measured admission rates into a
training corpus. Style, source composition and domain shift from the
validation material remain alternative explanations
(\Cref{fig:content-format}).

\subsection{Closed-model comparisons use different protocols}
The API runs recall 60.2\% of brands on Claude Haiku 4.5, 73.3\% on Gemini
3.5 Flash Lite and 78.5\% on GPT-5.6 Luna. These runs use four completions
per category, compared with twelve for Gemma. Their category tasks also
show candidates explicitly and permit an ``I do not know'' response,
whereas the Gemma probe ranks candidate-sentence likelihoods. These
differences prevent a controlled comparison of knowledge, error rates or
abstention policy. The API results are supplementary observations, not a
validation of the open-weight measurements.

\section{Discussion}
\label{sec:discussion}

The main finding is a distinction between being named in sampled category
completions and ranking a designated category correctly. Neither outcome
alone is an adequate summary of brand knowledge. They also have different
associations with the measured coverage variables, although the adjustment
sets differ and the analysis does not identify causal interventions.

The category margin offers a candidate diagnostic: larger margins coincide
with fewer errors in this cohort. Establishing its practical value requires
held-out calibration, sensitivity to alternative distractors and a test of
whether the relationship transfers to generated descriptions. A likelihood
margin should not be presented as spontaneous confidence or a calibrated
probability of correctness.

The commercial-fact and sentiment results further limit interpretation.
Weak discrimination on pricing-related probes does not prove that all
pricing answers require retrieval. A near-zero correlation with customer
ratings does not establish what the sentiment directions measure. Both
findings motivate validation against alternative prompts and independent
ground truth before these scores are used to guide decisions.

For future work, the most informative additions are a second open-weight
family, repeated sampling with uncertainty estimates, and a prospective
design that measures changes in coverage before later model releases.
These would test whether the reported patterns generalise and whether
changes in public information precede changes in either outcome.

\section{Limitations}
\label{sec:limitations}

\begin{itemize}\itemsep4pt

\item \textbf{Scope and sampling.} The convenience cohort contains English-language
B2B SaaS brands and is not representative of all companies. The open-weight
measurements cover one model family. Recall depends on prompts, sampling
depth, random seed and name matching; non-recall does not establish absence
from training.

\item \textbf{Probe validity.} A single designated category can mislabel a
multi-category company. Distractor choice and wording can alter likelihood
rankings even after length matching and padding corrections. The margin
analysis is retrospective, and the probes do not measure spontaneous
abstention or confidence. Closed-model tasks are not matched to the Gemma
protocol.

\item \textbf{Association analysis.} Coverage is measured observationally and
may reflect unmeasured company characteristics. Different regressions use
different complete-case samples. The owned-properties composite includes
crawl prominence, and the prominence baselines are unadjusted, so all rows
cannot be interpreted as effects under a common control set. Multiple-testing
correction does not resolve confounding, measurement error or exploratory
selection.

\item \textbf{Sentiment and membership inference.} Sentiment directions depend
on their prompt frames and overlap statistically. Their validity as measures
of real-world reputation is unestablished. Membership-detector calibration
on known documents does not prove membership of the surveyed passages or
exclude effects of style and distribution shift.

\item \textbf{Reproducibility.} The accompanying local artifacts document the
analysis, but the manuscript does not yet provide a public, versioned data
and code archive or a complete set of immutable model-checkpoint revisions.
These gaps limit independent replication of the reported runs.

\end{itemize}

\section{Conclusion}
\label{sec:conclusion}

In this 1,000-brand cohort, Gemma 4 12B recalls 51.0\% of brands under the
sampled category protocol and ranks the designated category first for
78.7\%. Larger category margins coincide with fewer errors, and recall and
category accuracy have different associations with public coverage. These
results support measuring the two outcomes separately. They do not establish
training-data membership, causal marketing effects or general performance
across model families.

\paragraph{Use of language-model tools.}
Claude and Codex assisted with manuscript preparation and revision. Their use
does not constitute independent validation of the experiments or findings.

\bibliography{references}

\begin{thebibliography}{7}
\providecommand{\natexlab}[1]{#1}
\providecommand{\url}[1]{\texttt{#1}}
\expandafter\ifx\csname urlstyle\endcsname\relax
  \providecommand{\doi}[1]{doi: #1}\else
  \providecommand{\doi}{doi: \begingroup \urlstyle{rm}\Url}\fi

\bibitem[Benjamini and Hochberg(1995)]{benjamini1995fdr}
Yoav Benjamini and Yosef Hochberg.
\newblock Controlling the false discovery rate: A practical and powerful
  approach to multiple testing.
\newblock \emph{Journal of the Royal Statistical Society: Series B},
  57\penalty0 (1):\penalty0 289--300, 1995.

\bibitem[Carlini et~al.(2021)Carlini, Tramer, Wallace, Jagielski, Herbert-Voss,
  Lee, Roberts, Brown, Song, Erlingsson, Oprea, and
  Raffel]{carlini2021extracting}
Nicholas Carlini, Florian Tramer, Eric Wallace, Matthew Jagielski, Ariel
  Herbert-Voss, Katherine Lee, Adam Roberts, Tom Brown, Dawn Song, Ulfar
  Erlingsson, Alina Oprea, and Colin Raffel.
\newblock Extracting training data from large language models.
\newblock In \emph{30th USENIX Security Symposium}, 2021.

\bibitem[Meng et~al.(2022)Meng, Bau, Andonian, and Belinkov]{meng2022rome}
Kevin Meng, David Bau, Alex Andonian, and Yonatan Belinkov.
\newblock Locating and editing factual associations in {GPT}.
\newblock In \emph{Advances in Neural Information Processing Systems
  (NeurIPS)}, 2022.

\bibitem[Petroni et~al.(2019)Petroni, Rockt{\"a}schel, Lewis, Bakhtin, Wu,
  Miller, and Riedel]{petroni2019lama}
Fabio Petroni, Tim Rockt{\"a}schel, Patrick Lewis, Anton Bakhtin, Yuxiang Wu,
  Alexander~H. Miller, and Sebastian Riedel.
\newblock Language models as knowledge bases?
\newblock In \emph{Proceedings of the 2019 Conference on Empirical Methods in
  Natural Language Processing (EMNLP)}, 2019.

\bibitem[Shi et~al.(2024)Shi, Ajith, Xia, Huang, Liu, Blevins, Chen, and
  Zettlemoyer]{shi2024mink}
Weijia Shi, Anirudh Ajith, Mengzhou Xia, Yangsibo Huang, Daogao Liu, Terra
  Blevins, Danqi Chen, and Luke Zettlemoyer.
\newblock Detecting pretraining data from large language models.
\newblock In \emph{International Conference on Learning Representations
  (ICLR)}, 2024.

\bibitem[Zhang et~al.(2025)Zhang, Sun, Yeats, Ouyang, Kuo, Zhang, Yang, and
  Li]{zhang2025minkpp}
Jingyang Zhang, Jingwei Sun, Eric Yeats, Yang Ouyang, Martin Kuo, Jianyi Zhang,
  Hao~Frank Yang, and Hai Li.
\newblock Min-k\%++: Improved baseline for detecting pre-training data from
  large language models.
\newblock In \emph{International Conference on Learning Representations
  (ICLR)}, 2025.

\bibitem[Zou et~al.(2023)Zou, Phan, Chen, Campbell, Guo, Ren, Pan, Yin,
  Mazeika, Dombrowski, Goel, Li, Byun, Wang, Mallen, Basart, Koyejo, Song,
  Fredrikson, Kolter, and Hendrycks]{zou2023representation}
Andy Zou, Long Phan, Sarah Chen, James Campbell, Phillip Guo, Richard Ren,
  Alexander Pan, Xuwang Yin, Mantas Mazeika, Ann-Kathrin Dombrowski, Shashwat
  Goel, Nathaniel Li, Michael~J. Byun, Zifan Wang, Alex Mallen, Steven Basart,
  Sanmi Koyejo, Dawn Song, Matt Fredrikson, J.~Zico Kolter, and Dan Hendrycks.
\newblock Representation engineering: A top-down approach to ai transparency,
  2023.
\newblock URL \url{https://arxiv.org/abs/2310.01405}.

\end{thebibliography}

\onecolumn
\appendix
\small
\section{Per-variable associations}
\label{app:adjusted}

\Cref{tab:adjusted} reports variables with adjusted $p<0.05$ in at least one
outcome-specific family. Names follow the analysis artifacts:
\texttt{explains\_<host>} is the share of brand snippets stating its category,
\texttt{count\_<host>} is mention volume, and \texttt{followers\_<host>} is
audience size. These exploratory associations are not intervention effects.
\begin{figure}[H]
\centering
\includegraphics[width=\textwidth]{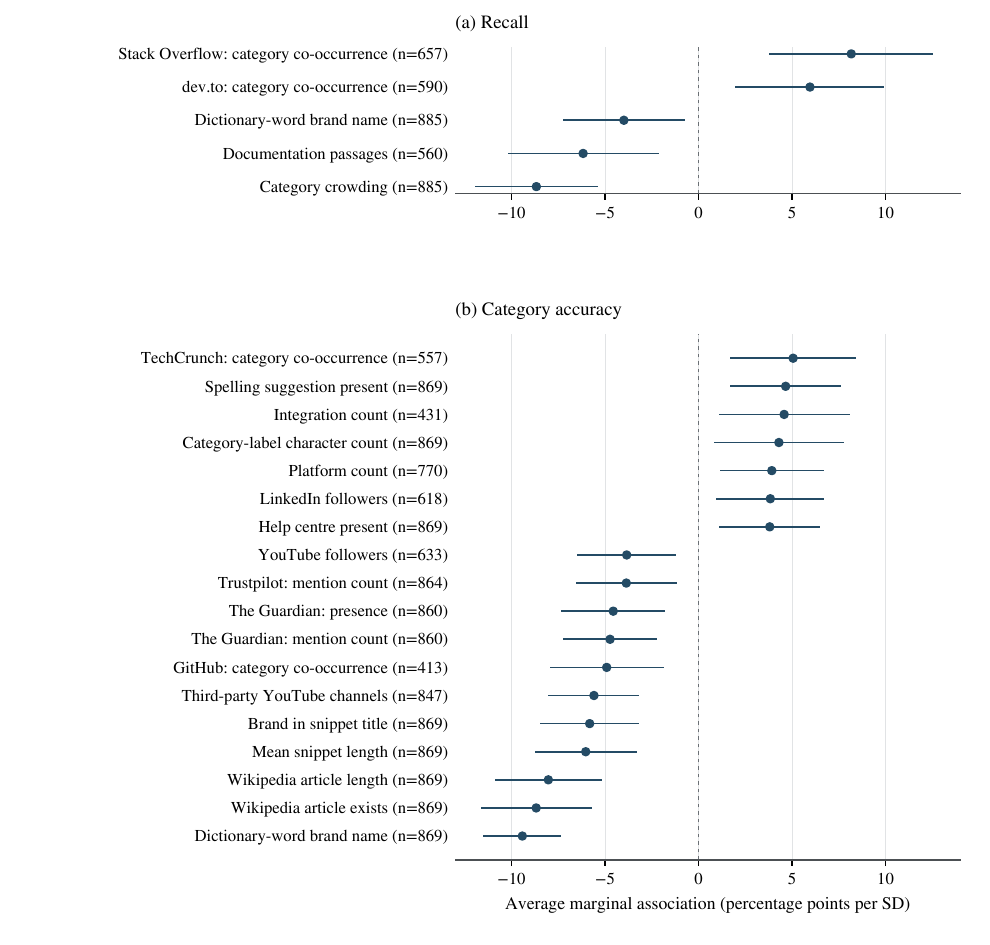}
\caption{Per-variable associations with within-family adjusted $p<0.05$, shown separately for recall and category accuracy. Points are percentage-point changes per predictor standard deviation; bars are approximate 95\% intervals. Selection is by corrected significance within each outcome and variable family, not by effect size. All models adjust for crawl rank and referring domains; accuracy also adjusts for reference-label word-count rank in the original candidate set.}
\label{fig:variable-associations}
\end{figure}

\begin{table}[H]
\centering
\caption{Variables with adjusted $p<0.05$ within family and outcome. Effects are percentage points per predictor standard deviation, controlling for crawl rank and referring domains, plus reference-label word-count rank in the original candidate set for accuracy. Intervals use scaled coefficient limits. All tested fits, including non-significant results, remain in the saved analysis artifact.}
\label{tab:adjusted}
\begin{tabular}{llrrrr}
\toprule
Variable & Family & $n$ & pts/SD & 95\% CI & $p_{adj}$ \\
\midrule
\multicolumn{6}{l}{\textit{Being named (recall)}} \\
category\_crowding & namespace & 885 & -8.67 & [-11.93, -5.42] & $<$0.001 \\
explains\_stackoverflow.com & coverage\_language & 657 & 8.17 & [3.83, 12.51] & 0.004 \\
docs\_passages & own\_property & 560 & -6.17 & [-10.16, -2.18] & 0.039 \\
explains\_dev.to & coverage\_language & 590 & 5.96 & [2.01, 9.90] & 0.028 \\
name\_is\_word & namespace & 885 & -3.99 & [-7.20, -0.79] & 0.037 \\
\midrule
\multicolumn{6}{l}{\textit{Being described correctly (category accuracy)}} \\
name\_is\_word & namespace & 869 & -9.42 & [-11.47, -7.38] & $<$0.001 \\
wiki\_article\_exists & own\_property & 869 & -8.68 & [-11.61, -5.75] & $<$0.001 \\
wiki\_article\_length & own\_property & 869 & -8.03 & [-10.83, -5.23] & $<$0.001 \\
mean\_snippet\_words & coverage\_volume & 869 & -6.03 & [-8.72, -3.33] & $<$0.001 \\
title\_share & coverage\_volume & 869 & -5.82 & [-8.44, -3.20] & $<$0.001 \\
yt\_third\_party\_channels & coverage\_volume & 847 & -5.59 & [-7.98, -3.20] & $<$0.001 \\
explains\_techcrunch.com & coverage\_language & 557 & 5.06 & [1.73, 8.39] & 0.026 \\
explains\_github.com & coverage\_language & 413 & -4.91 & [-7.90, -1.91] & 0.024 \\
count\_theguardian.com & coverage\_volume & 860 & -4.73 & [-7.22, -2.25] & 0.002 \\
spell\_suggested & namespace & 869 & 4.66 & [1.72, 7.59] & 0.005 \\
n\_integrations & own\_property & 431 & 4.58 & [1.13, 8.04] & 0.030 \\
on\_theguardian.com & coverage\_volume & 860 & -4.56 & [-7.30, -1.83] & 0.010 \\
category\_label\_chars & namespace & 869 & 4.30 & [0.89, 7.71] & 0.022 \\
n\_platforms & own\_property & 770 & 3.92 & [1.19, 6.65] & 0.020 \\
count\_trustpilot.com & coverage\_volume & 864 & -3.86 & [-6.53, -1.20] & 0.030 \\
followers\_youtube.com & coverage\_volume & 633 & -3.84 & [-6.45, -1.24] & 0.030 \\
li\_followers\_api & coverage\_volume & 618 & 3.84 & [1.00, 6.68] & 0.048 \\
has\_help\_centre & own\_property & 869 & 3.81 & [1.16, 6.46] & 0.020 \\
\bottomrule
\end{tabular}
\end{table}

\section{Numerical summaries}
\noindent\begin{minipage}[t]{0.48\textwidth}
\begin{table}[H]
\centering
\caption{Observed sampled recall across four Gemma checkpoints using the same cohort, prompt set and matcher. Differences are descriptive; model size and training also vary.}
\label{tab:generations}
\begin{tabular}{llr}
\toprule
Generation & Model & Recall \\
\midrule
Gemma 1 & gemma-7b       & 39.8\% \\
Gemma 2 & gemma-2-9b     & 44.3\% \\
Gemma 3 & gemma-3-12b-pt & 51.1\% \\
Gemma 4 & gemma-4-12B    & 51.0\% \\
\bottomrule
\end{tabular}
\end{table}

\end{minipage}\hfill
\begin{minipage}[t]{0.48\textwidth}
\begin{table}[H]
\centering
\caption{Category-ranking outcomes for 1,000 brands on Gemma 4 12B after label-length and padding corrections. Low separation is a probe outcome, not an explicit abstention.}
\label{tab:outcomes}
\small
\begin{tabular}{lrr}
\toprule
Outcome & Brands & Share \\
\midrule
Consistent (true category first)& 787 & 78.7\% \\
Deviated (a distractor first)    & 194 & 19.4\% \\
Low separation      &  19 &  1.9\% \\
\midrule
Median margin, consistent & \multicolumn{2}{r}{0.83} \\
Median margin, deviated   & \multicolumn{2}{r}{0.21} \\
\bottomrule
\end{tabular}
\end{table}

\end{minipage}

\noindent\begin{minipage}[t]{0.48\textwidth}
\begin{table}[H]
\centering
\caption{Observed error by winner-to-runner-up log-probability margin among 981 separated category rankings. These are in-sample results, not validated confidence probabilities.}
\label{tab:margin}
\begin{tabular}{lrr}
\toprule
Margin band & Brands & Error rate \\
\midrule
0.0--0.2 & 175 & 52.0\% \\
0.2--0.4 & 143 & 39.9\% \\
0.4--0.6 & 116 & 23.3\% \\
0.6--0.9 & 175 &  6.3\% \\
0.9--1.2 & 176 &  3.4\% \\
1.2--1.6 & 130 &  1.5\% \\
$\geq$1.6 &  66 &  0.0\% \\
\midrule
All above 0.6 & 547 & 3.5\% \\
\bottomrule
\end{tabular}
\end{table}

\end{minipage}\hfill
\begin{minipage}[t]{0.48\textwidth}
\begin{table}[H]
\centering
\caption{Commercial-fact discrimination using positive-minus-negative score contrasts. AUC has a chance baseline of 0.500. Category accuracy uses a different metric and baseline.}
\label{tab:facttypes}
\setlength{\tabcolsep}{3pt}
\small
\begin{tabular}{lrr}
\toprule
Fact type & Probes & Score \\
\midrule
Product category (5-way) & 1{,}000 & 78.7\%$^{\dagger}$ \\
Integration partner & 2{,}400 & 0.614 \\
Free tier exists        &   402 & 0.547 \\
Published pricing       &   585 & 0.526 \\
Self-serve signup       &   438 & 0.450 \\
\bottomrule
\multicolumn{3}{l}{\footnotesize $^{\dagger}$Accuracy on a 5-way rank against a 20\% chance}\\
\multicolumn{3}{l}{\footnotesize baseline, not on the AUC scale.}
\end{tabular}
\end{table}

\end{minipage}
\begin{table}[H]
\centering
\caption{Composite associations in outcome percentage points per predictor standard deviation. Benjamini--Hochberg correction covers 16 tests. Crawl and social prominence are unadjusted baselines. Owned properties includes crawl rank and adjusts only for referring domains. Other rows adjust for crawl rank and referring domains. Intervals use scaled coefficient limits as described in Methods.}
\label{tab:composites}
\begin{tabular}{lrrrrrrrr}
\toprule
& \multicolumn{4}{c}{Recall} & \multicolumn{4}{c}{Category accuracy} \\
\cmidrule(lr){2-5}\cmidrule(lr){6-9}
Composite & $n$ & pts/SD & 95\% CI & $p_{adj}$ & $n$ & pts/SD & 95\% CI & $p_{adj}$ \\
\midrule
Crawl prominence  & 885 & 12.58 & [9.21, 15.94] & $<$0.001 & 869 &  7.93 & [5.34, 10.53] & $<$0.001 \\
Social prominence & 909 &  9.75 & [5.22, 14.27] & $<$0.001 & 892 &  0.21 & [-2.43, 2.85] & 0.911 \\
Technical         & 885 &  5.37 & [1.99, 8.76]  & 0.006    & 869 & -2.77 & [-5.22, -0.32] & 0.047 \\
Explanation       & 885 &  3.92 & [0.70, 7.13]  & 0.045    & 869 &  3.15 & [0.47, 5.84]  & 0.047 \\
Review sites      & 884 &  3.87 & [0.39, 7.36]  & 0.047    & 868 & -0.80 & [-3.20, 1.60] & 0.586 \\
News              & 885 &  3.62 & [0.40, 6.83]  & 0.047    & 869 &  0.14 & [-2.35, 2.63] & 0.911 \\
Owned properties  & 998 &  1.36 & [-2.16, 4.87] & 0.574    & 979 &  5.92 & [2.79, 9.05]  & $<$0.001 \\
Social            & 871 &  1.29 & [-2.19, 4.77] & 0.574    & 855 & -1.83 & [-4.45, 0.79] & 0.248 \\
\bottomrule
\end{tabular}
\end{table}

\section{Reproduction and sensitivity}
\label{app:repro}

Distractors are sampled from the cohort categories under word-count matching,
using seed 20260828. Right-padded, attention-masked scoring agrees with
unpadded scoring. The saved runs show the following sensitivity:

\begin{center}
\small
\setlength{\tabcolsep}{3pt}
\begin{tabular}{lrrr}
\toprule
Run & Correct & Wrong & Unsep. \\
\midrule
Matched labels, right pad & 787 & 194 & 19 \\
Original labels, right pad & 753 & 225 & 22 \\
Matched labels, left pad & 663 & 320 & 17 \\
\bottomrule
\end{tabular}
\end{center}

Only the first run supplies the primary category results. The left-padded run
is invalid because the batch composition changes candidate scores. The earlier
substring recall matcher is also superseded by boundary-aware matching.

\begingroup\raggedright
The source artifacts are held under \texttt{experiments/landmark-study/} in
the study repository. Primary category outcomes are in
\path{results/public1k_facts_lenmatched_gemma4.json}; composite and per-variable
fits are in \path{results/composites_analysis.json} and
\path{results/adjusted_effects.json}. The scripts
\path{scripts/analyze_composites.py} and
\path{scripts/analyze_adjusted.py} specify the corresponding models.
Some descriptive metadata in the saved composite artifact predates the
addition of the two baselines; the eight fitted measures and their stored
control lists define the analysis reported here.
\par\endgroup

A public, versioned replication archive is not supplied with this manuscript.
Local artifact paths alone do not provide independent reproducibility. A
replication release should include the cohort and reference-label provenance,
exact prompts and matching rules, outcome thresholds, model checkpoint revisions,
sampling settings and analysis environment.

\end{document}